\documentclass[twoside,11pt]{article}

\usepackage[utf8]{inputenc}
\usepackage[T1]{fontenc}
\usepackage{bbm}
\usepackage{algorithmicx}
\usepackage{jmlr2e}
\usepackage{amsmath}
\usepackage{xcolor}
\usepackage{bm}
\usepackage{tikz}
\usetikzlibrary{bayesnet}
\usepackage{subcaption}
\usepackage{float}
\usepackage{stackengine}
\usepackage{array}
\usepackage{multirow}
\usepackage{stmaryrd}
\usepackage{pgfplots}
\usepgfplotslibrary{fillbetween}
\usetikzlibrary{patterns}
\usetikzlibrary{shapes.geometric}
\usepackage[ruled,vlined]{algorithm2e}

\pgfmathdeclarefunction{gauss}{3}{%
  \pgfmathparse{1/(#3*sqrt(2*pi))*exp(-((#1-#2)^2)/(2*#3^2))}%
}
\pgfmathdeclarefunction{sum_gauss}{5}{%
  \pgfmathparse{1/(3*#3*sqrt(2*pi))*exp(-((#1-#2)^2)/(2*#3^2)) + 2/(3*#5*sqrt(2*pi))*exp(-((#1-#4)^2)/(2*#5^2))}%
}

\definecolor{blue-violet}{rgb}{0.54, 0.17, 0.89}
\definecolor{bluepigment}{rgb}{0.2, 0.2, 0.6}
\makeatletter
\newcommand*\bigcdot{\mathpalette\bigcdot@{.5}}
\newcommand*\bigcdot@[2]{\mathbin{\vcenter{\hbox{\scalebox{#2}{$\m@th#1\bullet$}}}}}
\makeatother
\def\delequal{\mathrel{\ensurestackMath{\stackon[1pt]{=}{\scriptstyle\Delta}}}}
\definecolor{Yellow}{RGB}{211, 176, 15}
\definecolor{Green}{rgb}{0.01, 0.75, 0.24}
\colorlet{Red}{red!50!black}
\colorlet{Blue}{blue!50!black}
\definecolor{Violet}{rgb}{0.56,0.14,0.56}

\usepackage{setspace}
\usepackage[outdir=./]{epstopdf}
\usetikzlibrary{external}
\tikzset{external/system call={pdflatex \tikzexternalcheckshellescape 
	-halt-on-error
    -interaction=batchmode 
    -jobname "\image" "\texsource"
    && pdftops -eps "\image.pdf"}}
\jmlrheading{1}{2020}{1-48}{4/00}{10/00}{meila00a}{Théophile Champion and Marek Grze\'s and Howard Bowman}
\ShortHeadings{Branching Time Active Inference}{Champion et al.}
\firstpageno{1}

\begin{document}

\title{Branching Time Active Inference with Bayesian Filtering}

\author{\name Théophile Champion \email tmac3@kent.ac.uk \\
       \addr University of Kent, School of Computing\\
       Canterbury CT2 7NZ, United Kingdom
       \AND
       \name Marek Grze\'s \email m.grzes@kent.ac.uk \\
       \addr University of Kent, School of Computing\\
       Canterbury CT2 7NZ, United Kingdom
       \AND
       \name Howard Bowman \email H.Bowman@kent.ac.uk \\
       \addr University of Birmingham, School of Psychology,\\
       Birmingham B15 2TT, United Kingdom\\
       University of Kent, School of Computing\\
       Canterbury CT2 7NZ, United Kingdom
       }
       
\editor{\textbf{TO BE FILLED}} %TODO

\maketitle

\begin{abstract}% <- trailing '%' for backward compatibility of .sty file
Branching Time Active Inference \citep{AITS_THEORY,AITS_PRACTICE} is a framework proposing to look at planning as a form of Bayesian model expansion. Its root can be found in Active Inference \citep{FRISTON2016862,AI_TUTO,AI_VMP}, a neuroscientific framework widely used for brain modelling, as well as in Monte Carlo Tree Search \citep{6145622}, a method broadly applied in the Reinforcement Learning literature. Up to now, the inference of the latent variables was carried out by taking advantage of the flexibility offered by Variational Message Passing \citep{VMP_TUTO}, an iterative process that can be understood as sending messages along the edges of a factor graph \citep{FFG_TUTO}. In this paper, we harness the efficiency of an alternative method for inference called Bayesian Filtering \citep{BAYESIAN_FILTERING}, which does not require the iteration of the update equations until convergence of the Variational Free Energy. Instead, this scheme alternates between two phases: integration of evidence and prediction of future states. Both of those phases can be performed efficiently and this provides a seventy times speed up over the state-of-the-art.
\end{abstract}


\section{Introduction}

Active inference extends the free energy principle to generative models with actions \citep{FRISTON2016862,AI_TUTO,AI_VMP} and can be regarded as a form of planning as inference \citep{PAI}. Over the years, this framework has successfully explained a wide range of brain phenomena, such as habit formation \citep{FRISTON2016862}, Bayesian surprise \citep{bayes_surprise}, curiosity \citep{curiosity}, and dopaminergic discharge \citep{dopamine}. It has also been applied to a variety of tasks such as navigation in the Animal AI environment \citep{DeepAIwithMCMC}, robotic control \citep{pezzato2020active,sancaktar2020endtoend}, the mountain car problem \citep{catal2020learning}, the  game DOOM \citep{CULLEN2018809} and the cart pole problem \citep{cart_pole}.

However, because active inference defines the prior over policies as a joint distribution over the space of all possible policies, the method suffers from an exponential space and time complexity class. In the reinforcement learning literature, this problem can be tackled using Monte Carlo tree search (MCTS) \citep{6145622}, whose origins can be found in the multi-armed bandit problem \citep{Auer2002}. More recently, MCTS has been applied to a large number of tasks such as the game of Go \citep{Go}, the Animal AI environment \citep{DeepAIwithMCMC}, and many others.

More recently, Branching Time Active Inference (BTAI) \citep{AITS_THEORY,AITS_PRACTICE} proposed that planning is a form of Bayesian model expansion guided by the upper confidence bound for trees (UCT) criterion from the MCTS literature, i.e. a quantity from the multi-armed bandit problem whose objective is to minimize the agent's regret. And because the generative model is dynamically expanded, variational message passing (VMP) \citep{VMP_TUTO} was used to carry out inference over the latent variables. VMP can be understood as a flexible iterative process that sends messages along the edges of a factor graph \citep{FFG_TUTO}, and computes posterior beliefs by summing those messages together.

Bayesian filtering (BF) \citep{BAYESIAN_FILTERING} is an alternative inference method composed of two phases. In the first phase, Bayes theorem is used to compute posterior beliefs each time a new observation is obtained from the environment. In the second phase, posterior beliefs over the present state ($S_t$) are used to predict posterior beliefs over the state at the next time step ($S_{t+1}$). Importantly, this process is not iterative within a time step, i.e., it only contains a forward pass, and therefore is much more efficient than VMP.

In Section \ref{sec:ai_ts}, we present the theory underlying Branching Time Active Inference when using Bayesian filtering for inference over latent variables. In Section \ref{sec:results}, we show that using Bayesian filtering instead of variational message passing for the inference process provides BTAI with a seventy times speed-up while maintaining effective planning. Finally, Section \ref{sec:conclusion} concludes this paper and discusses avenues for future research.

\section{Branching Time Active Inference with Bayesian Filtering ($\text{BTAI}_{\text{BF}}$)} \label{sec:ai_ts}

\begin{figure}[H]
	\begin{center}
	\includegraphics[width=0.75\textwidth]{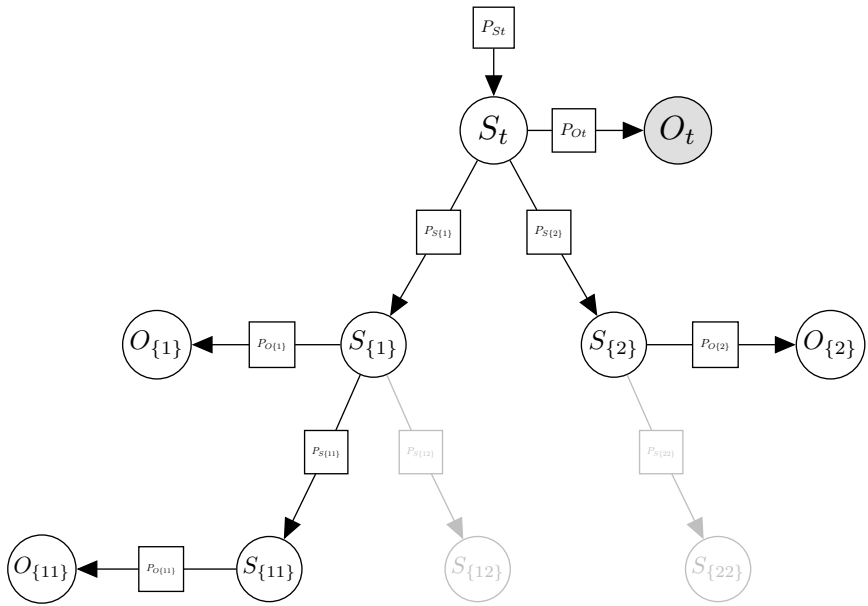}
 	\end{center}
\vspace{-0.25cm}
    \caption{
This figure illustrates the expandable generative model used by the BTAI with BF agent. The future is a tree like generative model whose branches correspond to the policies considered by the agent. The branches can be dynamically expanded during planning and the nodes in light gray represent possible expansions of the current generative model.}
    \label{fig:BTAI_BF}
\end{figure}

In this section, we describe the theory underlying our approach. For any notational uncertainty the reader is referred to Appendix F of \citet{AITS_THEORY}. We let $\bm{D}$ be a 1-tensor representing the prior over initial hidden states $P(S_0)$. Let $\bm{A}$ be a 2-tensor representing the likelihood mapping $P(O_\tau|S_\tau)$, and $\bm{B}$ be a 3-tensor representing the transition mapping $P(S_{\tau+1}|S_\tau, U_\tau)$. Additionally, we let $\mathbb{I}$ be the set of multi-indices containing all the policies (i.e., sequences of actions) that have been explored by the model. The generative model of BTAI with BF can be formally written as the following joint distribution:
\begin{align*}
P(O_{0},S_{0},O_{\mathbb{I}},S_{\mathbb{I}}) = &P(O_0|S_0) P(S_0) \prod_{I \in \mathbb{I}} P(O_I|S_I)P(S_I|S_{I \setminus \text{last}})
\end{align*}
where $S_{I \setminus \text{last}}$ is the parent of $S_I$, and:
\begin{align*}
&P(S_0) = \text{Cat}(\bm{D})& &P(O_\tau|S_\tau) = \text{Cat}(\bm{A})\\
&P(O_I|S_I) = \text{Cat}(\bm{A}) & &P(S_I|S_{I \setminus \text{last}}) = \text{Cat}(\bm{B}_I).
\end{align*}
where $\bm{B}_I = \bm{B}(\bigcdot, \bigcdot, I_{\text{last}})$ is the 2-tensor corresponding to $I_{last}$ (i.e., the last action that led to $S_I$), and the likelihood mapping in the past, i.e., $P(O_\tau|S_\tau)$, and in the future, i.e., $P(O_I|S_I)$, are both categorical distributions with parameters $\bm{A}$. This generative model is depicted in Figure \ref{fig:BTAI_BF}, where we assume that the current time step $t$ equals zero.

Initially, the generative model only contains the initial state $S_0$ and observation $O_0$. The prior over the hidden state is known, i.e. $P(S_0) = \text{Cat}(\bm{D})$, as well as the likelihood, i.e., $P(O_\tau|S_\tau) = \text{Cat}(\bm{A})$, and $P(O_0)$, the evidence, can be computed in the usual way by marginalizing over $P(O_0,S_0) = P(O_0|S_0)P(S_0)$. Thus, we can integrate the evidence provided to us by the initial observation $O_0$ using Bayes Theorem:
\begin{align}\label{eq:initial_beliefs}
\mathcal{B}(S_0) = \frac{P(O_0|S_0)P(S_0)}{P(O_0)},
\end{align}
where $\mathcal{B}(S_0)$ are the beliefs over the initial hidden state. Then, we use the UCT criterion to determine which node in the tree should be expanded. Let the tree's root $S_0$ be called the current node. If the current node has no children, then it is selected for expansion. Alternatively, the child with the highest UCT criterion becomes the new current node and the process is iterated until we reach a leaf node (i.e. a node from which no action has previously been selected). The UCT criterion \citep{6145622} for the $j$-th child of the current node is given by:
\begin{align}\label{eq:UCT}
UCT_j = - \bar{\bm{G}}_j + C_{explore} \sqrt{\frac{\ln n}{n_j}},
\end{align}
where $\bar{\bm{G}}_j$ is the average expected free energy calculated with respected to the actions selected from the $j$-th child, $C_{explore}$ is the exploration constant that modulates the amount of exploration at the tree level, $n$ is the number of times the current node has been visited, and $n_j$ is the number of times the $j$-th child has been visited.

Let $S_I$ be the (leaf) node selected by the above selection procedure. We then expand all the children of $S_I$, i.e., all the states of the form $S_{I::U}$ where $U \in \{1, ..., |U|\}$ is an arbitrary action, and $I::U$ is the multi-index obtained by appending the action $U$ at the end of the sequence defined by $I$. Next, we compute the predicted beliefs over those expanded hidden states using the transition mapping:
\begin{align}\label{eq:prediction}
\mathcal{B}(S_J) = \mathbb{E}_{\mathcal{B}(S_{I})}\big[P(S_J|S_I)\big],
\end{align}
where we let $J = I::U$ for any action $U$, $\mathcal{B}(S_{I})$ are the predicted posterior beliefs over $S_I$, and according to our generative model $P(S_J|S_I) = \text{Cat}(\bm{B}_J)$ with $\bm{B}_J = \bm{B}(\bigcdot, \bigcdot, J_{\text{last}})$. The above equation corresponds to the second phase of Bayesian filtering, i.e., the prediction phase, which involves	 the calculation of new beliefs, using the generative model, in the absence of new observations. Then, we need to estimate the cost of (virtually) taking each possible action. The cost in this paper is taken to be the expected free energy \citep{10.1162/NECO_a_00912}:
\begin{align}\label{eq:efe}
\bm{G}_J \delequal D_{\mathrm{KL}}[\mathcal{B}(O_J)||V(O_J)]\,\, +\,\, \mathbb{E}_{\mathcal{B}(S_J)}[\text{H}[P(O_J | S_J)]],
\end{align}
where the prior preferences over future observations are specified by the modeller as $V(O_J) = \text{Cat}(\bm{C})$, according to the generative model $P(O_J | S_J) = \text{Cat}(\bm{A})$, and the posterior beliefs over future observations are computed by prediction as follows:
\begin{align*}
\mathcal{B}(O_J) = \mathbb{E}_{\mathcal{B}(S_J)}[P(O_J | S_J)].
\end{align*}
Next, we assume that the agent will always perform the action with the lowest cost, and back-propagate the cost of the best (virtual) action toward the root of the tree. Formally, we write the update as follows:
\begin{align}\label{eq:backprop}
\forall K \in \mathbb{A}_I \cup \{I\}, \quad \bm{G}_K \leftarrow \bm{G}_K + \min_{U \in \{1, ..., |U|\}} \bm{G}_{I::U},
\end{align}
where $I$ is the multi-index of the node that was selected for (virtual) expansion, and $\mathbb{A}_I$ is the set of all multi-indices corresponding to ancestors of $S_I$. During the back propagation, we also update the number of visits as follows:
\begin{align}\label{eq:backprop_n}
\forall K \in \mathbb{A}_I \cup \{I\}, \quad n_K \leftarrow n_K + 1.
\end{align}
If we let $\bm{G}^{aggr}_K$ be the aggregated cost of an arbitrary node $S_K$ obtained by applying Equation \ref{eq:backprop} after each expansion, then we are now able to express $\bar{\bm{G}}_K$ formally as:
$$\bar{\bm{G}}_K = \frac{\bm{G}^{aggr}_K}{n_K}.$$
The planning procedure described above ends when the maximum number of planning iterations is reached, and the action corresponding to the root's child with the lowest average cost is performed in the environment. At this point, the agent receives a new observation $O_\tau$ and needs to update its beliefs over $S_\tau$. First, we predict the posterior beliefs over $S_\tau$ as follows:
\begin{align}\label{eq:empirical_prior}
\mathcal{B}(S_\tau|U_{\tau-1} = U^*) = \mathbb{E}_{\mathcal{B}(S_{\tau-1})}\big[P(S_\tau|S_{\tau-1},U_{\tau-1} = U^*)\big],
\end{align}
where $U^*$ is the action performed (from the root) in the environment, $P(S_\tau|S_{\tau-1},U_{\tau-1} = U^*)$ is the 2-tensor $\bm{B}(\bigcdot, \bigcdot, U^*)$, and $\mathcal{B}(S_{\tau-1})$ is the agent's posterior beliefs over the state at time $\tau-1$, e.g., after performing the first action in the environment, $\tau = 1$ and $\mathcal{B}(S_{\tau-1}) = \mathcal{B}(S_0)$ as given by Equation \ref{eq:initial_beliefs}. Second, we integrate the evidence provided by the new observation $O_\tau$ using Bayes theorem:
\begin{align}\label{eq:new_root_beliefs}
\mathcal{B}(S_\tau) = \frac{P(O_\tau|S_\tau)\mathcal{B}(S_\tau|U_{\tau-1} = U^*)}{P(O_\tau)},
\end{align}
where $\mathcal{B}(S_\tau|U_{\tau-1} = U^*)$ is used as an empirical prior. By an empirical prior we mean a posterior distribution of the previous time step, e.g., $\mathcal{B}(S_\tau|U_{\tau-1} = U^*)$, that is used as a prior in Bayes theorem. Algorithm \ref{algo:BTAI_BF_cycles} concludes this section by summarizing our approach.

\begin{algorithm}[H]
\label{algo:BTAI_BF_cycles}
\SetAlgoLined\DontPrintSemicolon
\SetKwInOut{Input}{Input}
\SetKwFor{RepTimes}{repeat}{times}{end}
\setstretch{1}
\SetAlgoLined
\Input{$env$ the environment, $O_0$ the initial observation, $\bm{A}$ the likelihood mapping, $\bm{B}$ the transition mapping, $\bm{C}$ the prior preferences, $\bm{D}$ the prior over initial states, $N$ the number of planning iterations, $M$ the number of action-perception cycles.}
 $\mathcal{B}(S_0) \leftarrow $ IntegrateEvidence($O_0$, $\bm{A}$, $\bm{D}$) \tcp*{Using (\ref{eq:initial_beliefs})}
 $root \leftarrow $ CreateTreeNode(beliefs = $\mathcal{B}(S_0)$, action = -1, cost = 0, visits = 1)\tcp*{Where -1 in the line above is a dummy value}
 \RepTimes{$M$} {
 \RepTimes{$N$} {
   $node \leftarrow $ SelectNode($root$) \tcp*{Using (\ref{eq:UCT}) recursively}
   $eNodes \leftarrow $ ExpandChildren($node$, $\bm{B}$) \tcp*{Using (\ref{eq:prediction}) for each action}
   Evaluate($eNodes$, $\bm{A}$, $\bm{C}$) \tcp*{Compute (\ref{eq:efe}) for each expanded node}
   Backpropagate($eNodes$) \tcp*{Using (\ref{eq:backprop}) and (\ref{eq:backprop_n})}
 }
 $U^* \leftarrow $ SelectAction($root$) \tcp*{Such that $U^*$ minimises the average cost}
 $O_\tau \leftarrow $ $env$.Execute($U^*$)\;
 $\mathcal{B}(S_{\tau-1}) \leftarrow root.beliefs$ \tcp*{Get beliefs of the root node}
 $\mathcal{B}(S_\tau|U_{\tau-1} = U^*) \leftarrow $ ComputeEmpiricalPrior($\bm{B}$, $\mathcal{B}(S_{\tau-1})$, $U^*$) \tcp*{Using (\ref{eq:empirical_prior})}
 $\mathcal{B}(S_\tau) \leftarrow $ IntegrateEvidence($O_\tau$, $\bm{A}$, $\mathcal{B}(S_\tau|U_{\tau-1} = U^*)$) \tcp*{Using (\ref{eq:new_root_beliefs})}
 $root \leftarrow $ CreateTreeNode(beliefs = $\mathcal{B}(S_\tau)$, action = $U^*$, cost = 0, visits = 1)\;
 }
 \caption{BTAI with BF: action-perception cycles (with relevant equations indicated in round brackets).}
\end{algorithm}

\section{Results} \label{sec:results}

In this section, we first present the deep reward environment in which two versions of BTAI will be compared. Then, we present experimental results comparing BTAI with VMP and BTAI with BF in terms of running time and performance.

\subsection{Deep reward environment}

This environment is called the deep reward environment because the agent needs to navigate a tree-like graph where the graph’s nodes correspond to the states of the system, and the agent needs to look deep into the future to diferentiate the good path from the traps. At the beginning of each trial, the agent is placed at the root of the tree, i.e., the initial state of the system. From the initial state, the agent can perform $n + m$ actions, where $n$ and $m$ are the number of good and bad paths, respectively. Additionally, at any point in time, the agent can make two observations: a pleasant one or an unpleasant one. The states of the good paths produce pleasant observations, while the states of the bad paths produce unpleasant ones.

If the first action selected was one of the $m$ bad actions, then the agent will enter a bad path in which $n + m$ actions are available at each time step but all of them produce unpleasant observations. If the first action selected was one of the $n$ good actions, then the agent will enter the associated good path. We let $L_k$ be the length of the $k$-th good path. Once the agent is engaged on the $k$-th path, there are still $n + m$ actions available but only one of them keeps the agent on the good path. All the other actions will produce unpleasant observations, i.e., the agent will enter a bad path.

This process will continue until the agent reaches the end of the $k$-th path, which is determined by the path's length $L_k$. If the $k$-th path was the longest of the $n$ good paths, then the agent will from now on only receive pleasant observations independently of the action performed. If the $k$-th path was not the longest path, then independently of the action performed the agent will enter a bad path.

\begin{figure}[H]
	\begin{center}
	\includegraphics[width=0.75\textwidth]{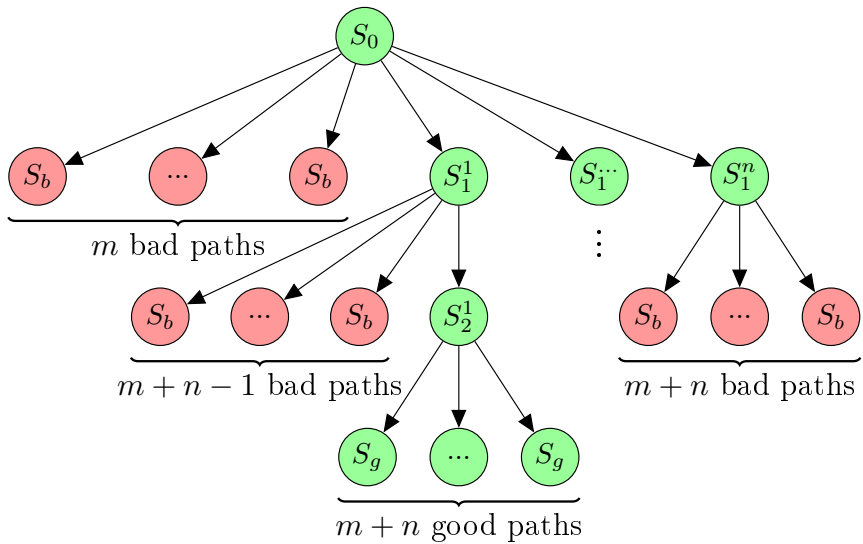}
 	\end{center}
\vspace{-0.25cm}
    \caption{
This figure illustrates a type of deep reward environment where $S_0$ represents the initial state, $S_b$ represents a bad state, $S_g$ represents a good state, and $S^i_j$ is the $j$-th state of the $i$-th good path. Also, the longest path in the above picture is the first good path whose length $L_1$ is equal to two. Importantly, the longest path corresponds to the only good path that does not turn out to be a trap.}
    \label{fig:graph_env}
\end{figure}

To summarize, at the beginning of each trial, the agent is prompted with $n$ good paths and $m$ bad paths. Only the longest good path will be beneficial in the long term, the others are traps, which will ultimately lead the agent to a bad state. Figure \ref{fig:graph_env} illustrates this environment.

\subsection{BTAI with VMP versus BTAI with BF}

In this section, we compare BTAI with VMP and BTAI with BF in terms of running time and performance. The running time reported in Tables \ref{tab:1} and \ref{tab:2} was obtained by running 100 trials each composed of 20 action-perception cycles. Also, the trial was stopped whenever the agent reached a bad state or the goal state. As shown in Tables \ref{tab:1} and \ref{tab:2}, both approaches were able to solve the tasks. However, BTAI with BF ran $69 \pm 26$ times faster than BTAI with VMP.

This speed up is possible for two reasons. First, Bayesian filtering does not require the iteration of the belief updates until convergence of the variational free energy. Second, when computing the optimal posterior over a random variable $X$, VMP needs to compute one message for each adjacent variable of $X$, add them together, and normalise using a softmax function. In contrast, BF only performs a forward pass, which is essentially implemented as matrix multiplications. 

\begin{table}[H]
\centering
\begin{tabular}{ |c|c|c|c|c|c|c| }
 \hline
 $n$ & $m$ & $L_1$, $L_2$, ..., $L_n$  & \# planning iterations & P(goal) & P(bad) & Running time\\
 \hline
 2 & 5 & 5, 8 & 25 & 1 & 0 & 2.294 sec\\
 \hline
 2 & 5 & 5, 8 & 50 & 1 & 0 & 4.688 sec\\
 \hline
 2 & 5 & 5, 8 & 100 & 1 & 0 & 9.045 sec\\
 \hline
 3 & 5 & 6, 5, 8 & 25 & 1 & 0 & 2.805 sec\\
 \hline
 3 & 5 & 6, 5, 8 & 50 & 1 & 0 & 5.416 sec\\
 \hline
 3 & 5 & 6, 5, 8 & 100 & 1 & 0 & 11.288 sec\\
 \hline
\end{tabular}
\caption{This table presents the results of BTAI with BF on various deep reward environments. Recall, that $n$ and $m$ are the number of good and bad paths, respectively. $L_i$ is the length of the $i$-th good path. $P(goal)$ reports the probability of reaching the goal state (i.e., the agent successfully picked the longest path), and $P(bad)$ reports the probability of reaching the bad state (i.e., either by picking a bad action directly of by falling into a trap).}
\label{tab:1}
\end{table}

\begin{table}[H]
\centering
\begin{tabular}{ |c|c|c|c|c|c|c| }
 \hline
 $n$ & $m$ & $L_1$, $L_2$, ..., $L_n$  & \# planning iterations & P(goal) & P(bad) & Running time\\
 \hline
 2 & 5 & 5, 8 & 25 & 1 & 0 & 4 min 16 sec\\
 \hline
 2 & 5 & 5, 8 & 50 & 1 & 0 & 4 min 42 sec\\
 \hline
 2 & 5 & 5, 8 & 100 & 1 & 0 & 6 min 26 sec\\
 \hline
 3 & 5 & 6, 5, 8 & 25 & 1 & 0 & 4 min 31 sec\\
 \hline
 3 & 5 & 6, 5, 8 & 50 & 1 & 0 & 5 min 35 sec\\
 \hline
 3 & 5 & 6, 5, 8 & 100 & 1 & 0 & 7 min 48 sec\\
 \hline
\end{tabular}
\caption{This table presents the results of BTAI with VMP on various deep reward environments. Recall, that $n$ and $m$ are the number of good and bad paths, respectively. $L_i$ is the length of the $i$-th good path. $P(goal)$ reports the probability of reaching the goal state (i.e., the agent successfully picked the longest path), and $P(bad)$ reports the probability of reaching the bad state (i.e., either by picking a bad action directly of by falling into a trap).}
\label{tab:2}
\end{table}

\section{Conclusion and future works} \label{sec:conclusion}

In this paper, we proposed a new implementation of Branching Time Active Inference \citep{AITS_THEORY,AITS_PRACTICE}, where the inference is carried out using Bayesian filtering \citep{BAYESIAN_FILTERING}, instead of using variational message passing \citep{AI_VMP,VMP_TUTO}.

This new approach has a few advantages. First, it achieves the same performance as its predecessor around seventy times faster. Second, the implementation is simpler and less data structures need to be stored in memory.

Also, one could argue that there is a trade-off in the nature and extent of the information inferred by classic active inference, branching-time active inference with variational message passing ($\text{BTAI}_{\text{VMP}}$) from \citet{AITS_THEORY,AITS_PRACTICE}, and branching-time active inference with Bayesian Filtering ($\text{BTAI}_{\text{BF}}$). Specifically, classic active inference exhaustively represents and updates all possible policies, while $\text{BTAI}_{\text{VMP}}$ will typically only represent one policly in the past (i.e., the one undertaken by the agent) and a small subset of the possible (future) trajectories. These will typically be the more advantageous paths for the agent to pursue, with the less beneficial paths not represented at all. Indeed, the tree search is based on the expected free energy that favors policies that maximize information gain, while realizing the prior preferences of the agent. $\text{BTAI}_{\text{BF}}$ stores even less data than $\text{BTAI}_{\text{VMP}}$, because the sequence of past hidden states is discarded as time passes, and only the beliefs over the current and future states are stored.

Additionally, full variational inference can update the system's understanding of past contingencies on the basis of new observations. As a result, the system can obtain more refined information about previous decisions, perhaps re-evaluating the optimality of these past decisions. Because classic active inference represents a larger space of policies, this re-evaluation could apply to more policies. When using Bayesian filtering, beliefs about past hidden states are discarded as time progresses, which makes Bayesian belief updating (about past hidden states) impossible.

We also know that humans engage in counterfactual reasoning \citep{rafetseder2013counterfactual}, which, in our planning context, could involve the entertainment and evaluation of alternative (non-selected) sequences of decisions. It may be that, because of the more exhaustive representation of possible trajectories, classic active inference can more efficiently engage in counterfactual reasoning. In contrast, branching-time active inference would require these alternative pasts to be generated ``a fresh'' for each counterfactual deliberation. In this sense, one might argue that there is a trade-off: branching-time active inference provides considerably more efficient planning to attain current goals, classic active inference provides a more exhaustive assessment of paths not taken. In contrast, branching time active inference implemented with Bayesian filtering does not leave a memory at all, let alone one upon which conterfactual reasoning could be realized.

The implementation of Branching Time Active Inference with variational message passing can be found here: $https://github.com/ChampiB/Homing-Pigeon$, and the implementation of Branching Time Active Inference with Bayesian Filtering is available on Github: $https://github.com/ChampiB/Branching\_Time\_Active\_Inference$.

Even with this seventy times speed up, BTAI is still unable to deal with large scale observations such as images. Adding deep neural networks to approximate the likelihood mapping is therefore a compelling direction for future research.

Also, this framework is currently limited to discrete action and state spaces. Designing a continuous extension of BTAI would enable its application to a wider range of applications such as robotic control with continuous actions.

Finally, as the depth of the tree increases, the beliefs about future states tend to become more and more uncertain, which can lead to a drop in performance. This suggests that there exists an optimal number of planning iterations, after which the model simply does not have enough information to keep planning. Future work could thus focus on automatically identifying this optimal number of planning iterations, in order to improve the robustness of the approach.

% Acknowledgements should go at the end, before appendices and references
\acks{TO BE FILLED}

\vskip 0.2in
\bibliography{references}

\end{document}